\pdfoutput=1

\documentclass[11pt]{article}

\usepackage[final]{acl}

\usepackage{times}
\usepackage{latexsym}
\usepackage{multirow}
\usepackage{colortbl}

\usepackage[T1]{fontenc}

\usepackage[utf8]{inputenc}

\usepackage{microtype}

\usepackage{inconsolata}

\usepackage{graphicx}

\usepackage[utf8]{inputenc} 
\usepackage[T1]{fontenc}    
\usepackage{hyperref}       
\usepackage{url}            
\usepackage{booktabs}       
\usepackage{amsfonts}       
\usepackage{nicefrac}       
\usepackage{microtype}      
\usepackage{xcolor}         
\usepackage[most]{tcolorbox}

\usepackage{multirow}
\usepackage[normalem]{ulem}
\useunder{\uline}{\ul}{}
\usepackage{tabularx}

\definecolor{darkgreen}{rgb}{0.0, 0.5, 0.0}
\usepackage{listings}
\definecolor{weizhey}{rgb}{0.9, 0, 0.1}

\definecolor{jing}{rgb}{0.1, 0.7, 0.1}
\newcommand{\jing}[1]{\textcolor{jing}{\bf\small [#1 --jing]}}
\renewcommand\sup[1]{$^{#1}$}

\definecolor{ping}{HTML}{9FBA95}

\renewcommand\sup[1]{$^{#1}$}

\usepackage{algorithm}
\usepackage{algorithmic}

\title{
Following Length Constraints in Instructions
}

\author{Weizhe Yuan\sup{1,2}~~~~~~~ Ilia Kulikov\sup{1}~~~~~~~  Ping Yu\sup{1}~~~~~~~ Kyunghyun Cho\sup{2}\\ \textbf{Sainbayar Sukhbaatar\sup{1}~~~~~~~ Jason Weston\sup{1,2}~~~~~~~ Jing Xu\sup{1}}\\
\\
\textsuperscript{1}Meta FAIR~~~~~~~~~~~~~~~
\textsuperscript{2}New York University~~~~~ }

\begin{document}
\maketitle
\begin{abstract}
Aligned instruction following models can better fulfill user requests than their unaligned counterparts. However, it has been shown that there is a length bias in evaluation of such models, and that training algorithms tend to exploit this bias by learning longer responses.
In this work we show how to train models that can be controlled at inference time with instructions containing desired length constraints.
Such models are superior in length instructed evaluations, 
outperforming standard instruction following models such as GPT4, Llama 3 and Mixtral.
\end{abstract}

\section{Introduction} \label{sec:intro}

Instruction following has emerged as one of the most 
important topics in AI, 
where the standard approach is to train
instruction-tuned large language models (LLMs)  
to respond to human requests \citep{ouyang2022training,touvron2023llama2}. 
One current challenge in developing better models is that there remain open questions on how to evaluate them, which in turn means there are open questions on how to train them with appropriate rewards.
It has been found that in current evaluations both humans and models tend to have a ``length bias'' whereby they prefer longer responses over shorter ones in pairwise preferences \citep{dubois2024alpacafarm}. Correspondingly, training methods that follow these preferences  tend to produce longer responses \citep{singhal2023long}.
Recently, instruction following benchmarks 
have incorporated length penalties into their scoring mechanisms to counteract this bias \citep{dubois2024length}, but  
this does not fix the problem at its source.

In this work, we argue that the expected length of responses is ill-defined in many queries, and this ambiguity makes evaluation difficult, which in turn affects training algorithms that use these evaluation signals. To resolve this we propose that evaluation should include {\em further disambiguating instructions} that prescribe the length of the desired response. 
 Typical requests can be ambiguous in terms of the length of the desired response, for example without context the instruction {\em `Give me information about Coco Gauff''} could be answered by a few sentences, a few paragraphs, or a multi-page document.
Yet, given the context, the intended length is often clearer, for example the expected length of the replies in the downstream application the user is interacting with, the interface being used (voice, viewed on a phone vs. a laptop) and so on. Hence adding a further instruction 
in a given context to the above example
such as {\em "The answer should be 300 words or less"} resolves this ambiguity.\footnote{We note that this paper itself was generated (by humans!) with the constraint that it has to be at most 8 pages.}

\begin{table*}[t!]
\centering 
\footnotesize
\renewcommand{\arraystretch}{1.2}
\begin{tabularx}{\textwidth}{m{2.3cm}>{\centering\arraybackslash}m{1.3cm}m{9.7cm}>{\centering\arraybackslash}X}
\toprule
\cellcolor{yellow!20}  \textbf{Instruction} &\cellcolor{green!20}\textbf{\texttt{<MAX\_LEN>}}&  \cellcolor{blue!20} \textbf{Generation} &  \cellcolor{red!20} \textbf{Length} \\
\midrule
\cellcolor{yellow!7} 
\cellcolor{yellow!7} & \cellcolor{green!7}  20  &  \cellcolor{blue!7} Yes, I'd be happy to help you improve your handwriting skills! &  \cellcolor{red!7}  12 \\
\cline{2-4}
\cellcolor{yellow!7} 
Answer the following instruction 
using \textbf{\texttt{<MAX\_LEN>}}  words or less.
& \cellcolor{green!7}  40 & \cellcolor{blue!7} I'd be happy to help. To improve your handwriting, practice regularly, focus on posture and grip, and try using guidelines or a light box to help you stay within lines. & \cellcolor{red!7}  31 \\
\cline{2-4}
\cellcolor{yellow!7} I'm trying to teach myself to have nicer handwriting.  Can you help? & \cellcolor{green!7}  80 & \cellcolor{blue!6}
I'd be happy to help you improve your handwriting. Here are a few tips to get you started: \textcolor{blue!6}{dslkfhhsadlfkadskfkladsjkfladskjfkladsjkf ff adsjk jadds d}*~Choose the right tools: Use a pen with good grip and smooth ink flow. Experiment with different types of paper to find one that works well for you.  & \cellcolor{red!7} 
 70 \\
\cellcolor{yellow!7}& \cellcolor{green!7}  & \cellcolor{blue!6}~~* Practice regularly: Set aside time each day to practice writing.& \cellcolor{red!7}  \\
\cellcolor{yellow!7} &\cellcolor{green!7}  &\cellcolor{blue!6}~~* Focus on posture and grip: Keep your hand steady and your fingers relaxed.& \cellcolor{red!7}  \\
\midrule

\multirow{2}{=}{
\begin{minipage}[t]{2.3cm} 
\cellcolor{yellow!10} I'm trying to teach 
\end{minipage}
} 

 \cellcolor{yellow!10}  & \cellcolor{green!7}  N/A & \cellcolor{blue!6} 
Improving your handwriting takes practice, patience, and some guidance. I'd be happy to help you with that. Here are some tips to get you started: & \cellcolor{red!7}  456
\\
\cellcolor{yellow!7} myself to have nicer handwriting. Can you help? & \cellcolor{green!7}  & \cellcolor{blue!6} 1. **Choose the right tools**: Invest in a few good quality pens or pencils that feel comfortable in your hand. Experiment with different types of pens (e.g., fine tip, medium tip, calligraphy) to find what works best for you. &\cellcolor{red!7}  \\
\cellcolor{yellow!10} &\cellcolor{green!7}  & \cellcolor{blue!6}  2. **Practice basic strokes**: Focus on mastering basic strokes such as: ...... & \cellcolor{red!7}  \\
\bottomrule
\end{tabularx}
\caption{{\bf Length-Instructed example generations.} We show examples from our 
Length-Instruction Fine-Tuned (LIFT)
Llama-3-8B-Instruct model with different length instruction limits for the same question. The last row is a response generation using  the original input without length instructions (partial generation due to limited space). Many  state-of-the-art LLMs are unable to follow such length instructions, see \autoref{fig:scatter_gpt4_llama3}.
\label{tab:the_example}
} 
\end{table*}

We show that many existing state-of-the-art instruction following models fail to follow such maximum word length instructions adequately. To measure this we construct and evaluate models on length instructed versions of AlpacaEval 2 \citep{dubois2024alpacafarm}  and MT-Bench \citep{zheng2023judging} by augmenting existing prompts with length instructions.  We find that, for example, GPT4-Turbo  violates length constraints almost 50\% of the time, highlighting a significant flaw in these models when it comes to steering their output length. 

We hence develop a method for improving instruction following models at length instruction following.  Our approach, Length-Instruction Fine-Tuning (LIFT),
involves taking a conventional instruction following dataset and constructing augmented training data by inserting length instructions in the original prompts.
We define length instructions so that the constructed preference pairs reflect both length constraints and response quality.
This length instruction augmented dataset is used in finetuning a model via Direct Preference Optimization (DPO) \citep{rafailov2023direct}.
We train both Llama 2 and Llama 3 models using LIFT-DPO 
and evaluate them on our length instructed benchmarks.
See \autoref{tab:the_example} for some example length instructed generations.
We find that our method leads to less length constraint violations and improved overall win rates
compared to existing instruction following models.

\section{Related Work}
\subsection{Length Bias in Model Alignment}

When optimizing for instruction following ability, reinforcement learning (RL) has been consistently observed to encourage models to produce longer responses \citep{singhal2023long}. 
\citet{zhao2024long} showed that simply selecting the longest data from the training set for fine-tuning is a strong baseline, and \citet{singhal2023long} showed that optimizing for response length  
is a significant factor behind RL’s reported improvements.
This effect seen in training parallels that on the evaluation side, whereby both humans and models 
tend to favor longer responses over shorter ones \citep{dubois2024alpacafarm}. 
Correspondingly, constructing preference pairs either through human feedback (RLHF) or through 
AI feedback (RLAIF) is likely to reflect these biases.
%
%
On the other hand, longer responses are not necessarily better even if preferred by annotators \citep{park2024disentangling}. For example they are more likely to contain inaccuracies \citep{achiam2023gpt}, which may be missed 
by human evaluators on challenging tasks \citep{casper2023open}.

Recently, instruction following benchmarks 
such as AlpacaEval 2 \citep{dubois2024length} and WildBench \citep{allenai2024wildbench} 
have incorporated length penalties into their scoring mechanisms to counteract this bias.
This is done by fitting a  generalized linear model to predict the (biased) preferences given length as a feature, and then obtaining length-debiased preferences by predicting preferences 
after removing the length term from the regression.
While this penalizes longer responses to a certain degree,
it is not yet clear if this new scoring function can still be gamed by models.


\subsection{Length-aware Model Training}

Learning methods that take into account length have historically been
prevalent in the task of summarization, see for example \citet{fan2017controllable}, and in particular 
\citet{goyal2022news,jie2023prompt} for length constrained summarization.

For instruction following,  \citet{singhal2023long} investigate several mitigations, such as balancing preferences or truncating lengths, but  do not find that they uniformly help.
 Both  \citet{shen2023loose} and \citet{chen2024odin}
propose to modify the reward model to disentangle length from quality so that they can concentrate the training on quality.
\citet{park2024disentangling} proposes modifying the DPO objective function with a length regularizer, and reports this prevents length exploitation while maintaining quality.
These approaches all assume there is an optimum length of responses which the model should be trained to generate. In contrast, our work assumes desired length depends on additional context, and a good model should be capable of following length instructions (i.e., via prompting for desired length).

\begin{figure}[t!]
\centering
\small
\begin{tcolorbox}[colback=green!10!white, 
                  colframe=green!30!white, 
                  arc=4mm, 
                  auto outer arc,
                  ]
Answer the following instruction using \texttt{\textcolor{red}{<MAX\_LEN>}} words or less.\\
\\
\texttt{\textcolor{red}{<ORIGINAL\_INSTRUCTION>}}

\end{tcolorbox}
\caption{{\bf Length Instruction Following.} We define the above prompt template in order to require models to produce responses within a maximum response length.}
\label{fig:lenctrl_template}
\end{figure}

Some production LLMs incorporate system prompts that reference output length, for example the sentence   ``give concise responses to very simple questions, but provide thorough responses to more complex and open-ended questions'' is included in the system prompt of Claude 3 \citep{Claude3}. However, to our knowledge
no systematic evaluation of such prompts has been reported.

Many {\em post-training} 
datasets used to fine-tune popular large language models have not been released or their exact makeup detailed 
\cite{achiam2023gpt,touvron2023llama2} hence it is difficult to immediately ascertain if such length following instructions are contained in their training setups. 
However, from data that has been released \citep{bai2022training,kopf2024openassistant} it appears that the amount of such data is generally small. 
Instead, preference pairs are typically provided which implicitly assume a preferred target length for a given prompt, while the prompt itself does not contain length following instructions. 
Such preferences are well known to typically prefer longer responses over shorter ones
 \citep{singhal2023long}.

\begin{figure*}[t!]
    \includegraphics[width=0.48\linewidth]{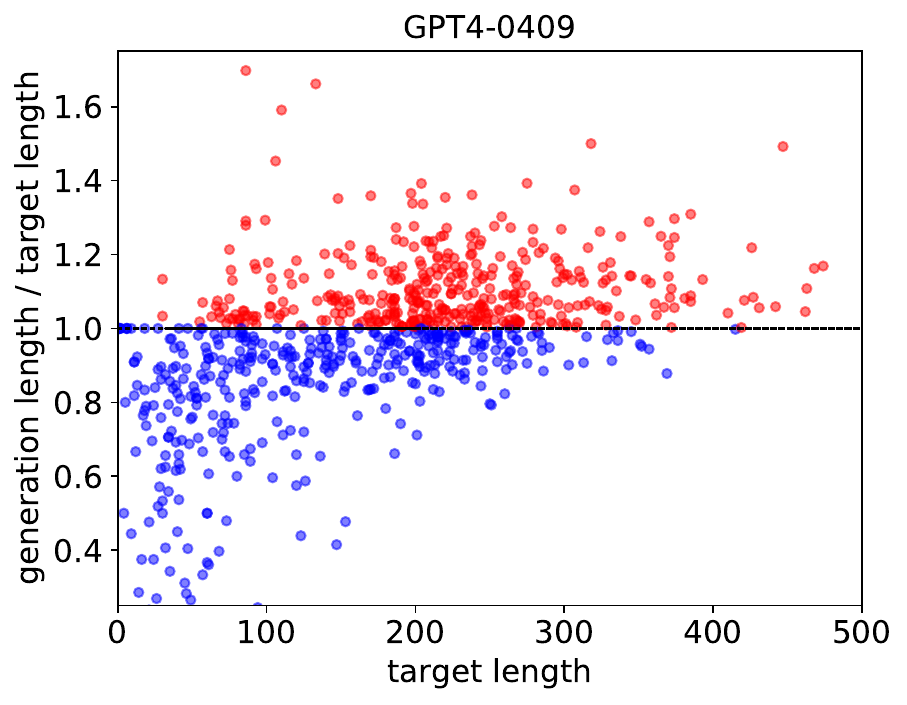}
    \includegraphics[width=0.48\linewidth]{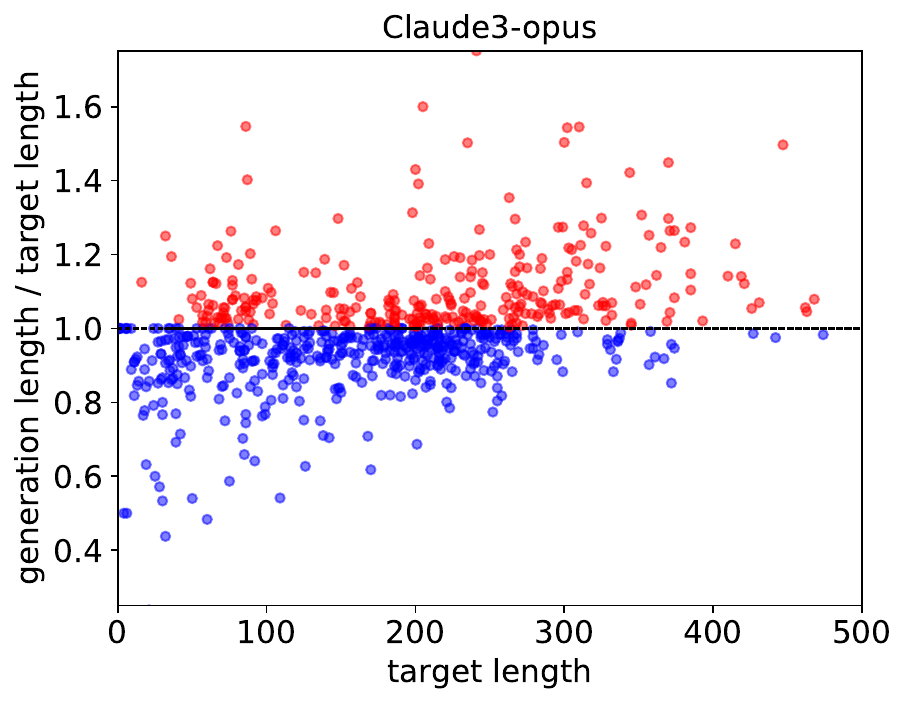}
    \hfill
    \caption{
    {\bf SOTA Models fail to follow length instructions}. Length instruction following of GPT4-0409 and Claude3-Opus  on 802 AlpacaEval-Length-Instructed (LI) examples. The  target length is plotted on the x-axis and the ratio of the actual generated length to the target length on the y-axis. Red dots represent violations where the generated length exceeds the target limit, while blue dots satisfy the limit.
    }
    \label{fig:scatter_gpt4_llama3}
\end{figure*}

\section{AlpacaEval-LI \& MT-Bench-LI: ~~~~~~New Length-Instructed Benchmarks}
Strongly performing instruction following models should naturally be able to follow given length limits, as those are instructions as well. Such instructions can be a natural part of a prompt, for example {\em "Tell me about <concept>. The answer should be 300 words or less"}. Depending on the use case and context, users might want responses from the same underlying model but of a different length, e.g. either a shorter or longer answer. 
In this section, we thus first evaluate the ability of current instruction following models to follow length instructions. In order to do this, we thus build
length-instructed (LI) benchmarks, AlpacaEval-LI and  MT-Bench-LI\footnote{Our  length-instructed benchmarks are available at 
\tiny\url{https://github.com/facebookresearch/RAM/tree/main/projects/length_instruct}}.

\subsection{Augmenting General Instructions with Length Constraints}
\label{sec:extend}
To evaluate a model's length instruction-following ability,
we augment existing instruction-following tasks by inserting maximum length limits as part of the instructions, as shown in the template in \autoref{fig:lenctrl_template}. 
This tests whether models can respond to the given query successfully, whilst also fulfilling the given length instruction.

\subsubsection{Target Length} 
\label{sec:target-choice}
The choice of desired length limits might vary a lot by instruction and task. To establish a reasonable yet challenging length limit for effectively evaluating current state-of-the-art (SOTA) models, we base the target length limit on the generation lengths of three strong SOTA models: GPT-4 Turbo (11/06) \citep{achiam2023gpt}, Claude 3 Opus (02/29)\footnote{https://www.anthropic.com/news/claude-3-family} and Mistral Large (24/02)\footnote{https://mistral.ai/news/mistral-large/}. 
We set {\em <MAX\_LEN> } in the template to the minimum generation length among these three models given the original prompts. 
Therefore, this length constraint varies for each individual prompt, and is short enough to be challenging, i.e., is not trivially satisfied by all SOTA models.

\subsubsection{Length Instruction-Following Baseline} 
\label{sec:baseline}
Many benchmarks evaluate models by conducting pairwise comparisons between model outputs. For given instructions, they report win rates against a strong baseline, such as GPT-4 generations using
another LLM to judge the pair (LLM-as-a-Judge).

To establish a strong 
baseline that consistently adheres to the length constraint, we employ the same minimum of three models approach from \autoref{sec:target-choice}.
Thus instead of a single model, each baseline response is chosen to be the shortest response generated from the three models.
This ensures that the baseline generations always meet the length constraint specified in the prompt while maintaining high generation quality. Thus, for each model tested we compare its generations 
with this baseline in a pairwise setting.

\subsubsection{Metrics}
We propose two metrics: Length-Instructed (LI) winrates against the baseline to evaluate {\em response quality} and violation rates to measure {\em length instruction following} ability.

\begin{figure*}[t!]
    \centering
    \includegraphics[width=1\linewidth]{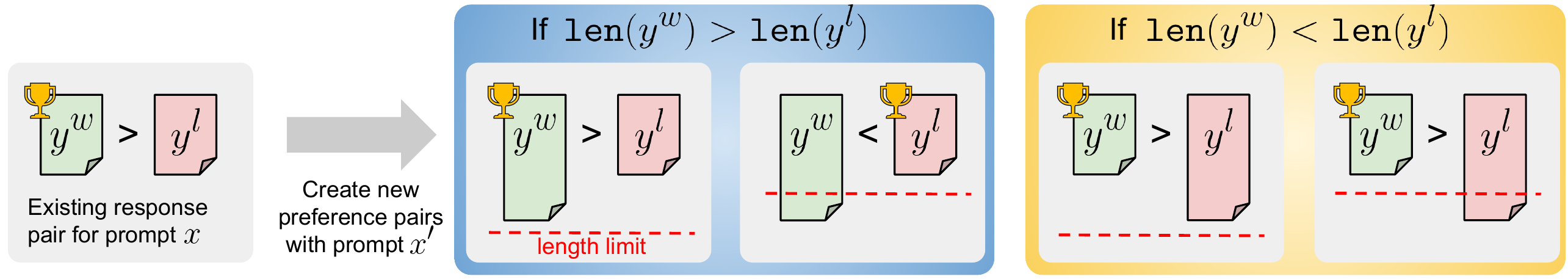}
    \caption{Length-Instruction Fine-Tuning (LIFT) method for augmenting preference pairs $(x, y^w, y^l)$ in general instruction-following tasks with length instructions. The original prompt $x$ is prepended with an extra instruction specifying a target response length, yielding a new prompt $x^\prime$. 
    The target length is chosen in multiple ways, creating extra preference pairs. In some cases the original winning response $y^w$ becomes the loser if it exceeds the limit.
    }
    \label{fig:li_data_construction}
\end{figure*}

\paragraph{Length Instruction Following}
We use violation rates (Vlt\%) to measure the percentage of responses that exceed the length constraint by counting the number of words. Additionally, we report other metrics, such as the average response length (in words). To calculate the word count of a response we use the word tokenization function provided by NLTK, excluding punctuation. The exact word count function is detailed in \autoref{sec:appex_wordcount}.

\paragraph{Response Quality}
We report  winrates from pairwise comparisons between model and baseline generations on  length-following instructions, referred to as the Length-Instructed (LI) Winrate.

The winner of each pairwise comparison is determined by both the quality of the responses and adherence to the length constraints. We treat the length limit as a hard constraint.
Since the baseline always satisfies the length constraint, if the model response being tested exceeds the limit it automatically loses. If the model response satisfies the length limit, we use the standard pairwise LLM-as-a-Judge comparison between the two
responses, where only the original instruction without length limit is given to the judge as input. 

\subsection{Length-Instructed AlpacaEval}
\label{sec:alpaca_leneval}

AlpacaEval 2 \citep{dubois2024alpacafarm} 
is an evaluation task
consisting of 805 general instruction following  prompts 
from 
creativity, brainstorming and writing to question answering, math and reasoning tasks.
We augment this task with length instructions to create the \emph{AlpacaEval-Length-Instructed (LI)} benchmark as described in \autoref{sec:extend}.

Following \autoref{sec:target-choice} we take the minimum generation length of the three strong LLMs as a target length for each prompt. Three out of the 805 Alpaca test instructions already have an explicit length constraint in the original prompt. We therefore only consider the remaining 802 prompts for the AlpacaEval-LI benchmark.

\autoref{fig:scatter_gpt4_llama3} shows the ratio of generation lengths over target instruction lengths as target lengths vary. GPT4-0409 generations exceed the target length limits almost 50\% of the time (red dots), especially when target lengths are over 200 words. Claude3-Opus has a similar trend 
according to the scatter plot. We also include results for Mistral Large and LLAMA3-70b-Instruct in \autoref{sec:appex_sota_len_following_claude3_mistral}.

Standard AlpacaEval 2 compares model outputs against baseline GPT-4 Turbo generations. In  AlpacaEval-LI, 
the baseline is built from
GPT4-1106, Claude3-Opus and Mistral Large as described in \autoref{sec:baseline}.
Their respective winrates in the standard AlpacaEval 2 are 50\%, 40.5\% and 32.7\%. This indicates that the resulting baseline is of high quality while consistently meeting the length constraint.

\subsection{Length-Instructed MT-Bench}
\label{sec:mtbench_leneval}
In addition, we also extend the MT-Bench evaluation \citep{zheng2023judging}  with length instructions to test models on a wide-range of prompts.
This dataset consists of 80 challenging multi-turn questions covering 8 categories of user prompts (writing, roleplay, extraction, reasoning, math, coding, STEM, and humanities/social science).

We follow the same steps as described in \autoref{sec:extend} on the MT-Bench evaluation set by sampling three length constraints for each prompt.
For simplicity we only consider first turns,  giving 240 MT-Bench-LI prompts. We will use this benchmark, along with AlpacaEval-LI in our experiments in
\autoref{sec:results}.

\section{Length-Instruction Fine-Tuning (LIFT)}
As shown in the previous section, current SOTA models may not adhere to specific length following instructions. To improve the ability of models in length-instruction following tasks, we propose the following method, which first builds 
Length-Instruction Fine-Tuning (LIFT) data.
This   training data consists of preference pairs, which can be used for training  models via RLHF or other preference optimization methods.

We first assume we are given an existing pairwise preference dataset $\mathcal{D}$ consisting of $N$ triples of input prompt, winning response,  and losing response $(x, y_i^w, y_i^l)_{i=1\cdots N}$. 
Let us denote by \texttt{len}$(y)$ the number of words in response $y$.
First, we filter out any triple where the difference between \texttt{len}($y_i^w$) and \texttt{len}($y_i^l$) is less than a certain threshold $T$ ($T=10$ in our experiments). 
We then construct an augmented dataset $\mathcal{D}^\prime$ that prepends an explicit length instruction to the input prompt $x_i$ using the template shown in \autoref{fig:lenctrl_template} to convert it into $x_i^\prime$.
We then construct new length-instructed preference pairs 
$(x_i^\prime, y_i^{w^{'}}, y_i^{l^{'}})$ 
where the winners and losers of the pairs
are determined as follows:


\begin{itemize}
    \item If $\texttt{len}(y_i^w) > \texttt{len}(y_i^l)$, i.e. the winning response is longer, we construct two samples in the augmented dataset $\mathcal{D}^\prime$ by, (1) adding a length instruction to $x_i$ that both responses satisfy (we simply use $\texttt{len}(y_i^w) + T$) and the winning response and losing response remain the same, and (2) adding a length constraint uniformly sampled from the interval $\left[\texttt{len}(y_i^l), \texttt{len}(y_i^w)\right]$, and $y_i^w$ becomes the losing one due to the violation of length constraint, and $y_i^l$  becomes the winning one.
    \item If $\texttt{len}(y_i^w) < \texttt{len}(y_i^l)$, we also construct two samples in the augmented dataset $\mathcal{D}^\prime$ by, (1) adding a length constraint to $x_i$ that both responses satisfy (we simply use $\texttt{len}(y_i^l) + T$), and (2) adding a length constraint sampled from the interval $\left[ \texttt{len}(y_i^w), \texttt{len}(y_i^l) \right]$. In both (1) and (2), the winning response and the losing response remain the same as in the original dataset.
\end{itemize}
The data construction process is also illustrated in \autoref{fig:li_data_construction}. Having these preferences will ensure models can handle a wide-range of target lengths and prioritize the length constraint over the original preferences when necessary. We use DPO to train our models, using both datasets $D$ and $D'$ so that models can handle prompts with and without length instructions.


\section{Experimental Setup}

We empirically investigate model performance on following length instructions, and the effectiveness of our LIFT training strategy. We begin with a description of our experimental setup.

\begin{table*}[t!]
 \small
\setlength{\tabcolsep}{11.5pt}
  \centering
\begin{tabular}{lrrrcccc}
\toprule
& \multicolumn{3}{c}{AlpacaEval-LI}  &   \multicolumn{3}{c}{MT-Bench-LI} \\
\cmidrule(lr){2-4} \cmidrule(lr){5-7}
\emph{Standard models}  & \multicolumn{1}{c}{Vlt(\%)}   & \multicolumn{1}{c}{Win(\%)} & \multicolumn{1}{c}{Words} & \multicolumn{1}{c}{Vlt(\%)}   & \multicolumn{1}{c}{Win(\%)} & \multicolumn{1}{c}{Words} \\
\midrule
{GPT4 Omni (gpt-4o-2024-05-13)}   
& 39.0  & 35.7 & 180 &  39.2 & 30.2 & 177 \\
{GPT4 Turbo (gpt4\_1106\_preview)} 
& 46.1 & 29.9 & 182 &  45.0 & 28.1 & 174 \\
{GPT4 Turbo (gpt-4-turbo-2024-04-09)}     
& 49.3  & 29.2  & 187 & 44.2 & 27.5 & 179 \\
{Claude 3 Opus (02/29)}  
&  37.0  &  32.9 & 183 & 37.9 & 33.1 & 174 \\
{Mistral Large (24/02)} 
& 17.6 & 28.8 & 158 & 20.8 & 27.7 & 158 \\
{Llama3-70B-Instruct}   & 10.2  & 38.5  &  154   &  20.3 & 28.5 & 151 \\
{Llama3-8B-Instruct}  
& 7.0 & 22.5 &  145 & 20.0 & 20.0 & 140 \\
{Llama2-70B-Chat}    
& 28.2 & 11.3  & 162 & 38.3 & 11.9 & 168\\
{Llama2-70B-Base (zero shot)}    
& 62.6 & 0.6  & 582& 63.8 & 1.7 & 293 \\
{Llama3-8B-Base (zero shot)}  
& 71.6 &  0.2 & 1935 & 27.9 & 2.3 & 185 \\
\bottomrule
\end{tabular}
\caption{{\bf  Length Instruction-Following results of SOTA models on the AlpacaEval-LI + MT-Bench-LI benchmarks.}
Many SOTA LLMs have large violation rates (Vlt(\%)) as they fail to follow length instructions.
}
  \label{tab:mt-bench-sota}
\end{table*}

\subsection{Train Dataset \& Baselines}
\paragraph{Standard Training Data}
We use the human-authored examples provided in the OpenAssistant (OA) dataset \citep{kopf2023openassistant} for instruction fine-tuning. Following \citet{li2023self} we use 3,200 examples as $\mathcal{D}$, by sampling only the first conversational turns in the English language that are high-quality, based on their human annotated rank (choosing only the highest rank 0 as chosen and rank 1 as loser). 
We first do supervised finetuning (SFT) on the chosen responses of $\mathcal{D}$.
We then further fine-tune the SFT model using the DPO loss on response pairs in $\mathcal{D}$, which becomes our \emph{Standard DPO} baseline.
In addition, we also compare against the Length Regularized DPO (R-DPO)  \cite{park2024disentangling} baseline that penalizes longer responses by modifying the DPO loss.

\paragraph{Length-Instructed  Fine-Tuning (LIFT) Data}
We apply our LIFT method to create dataset $\mathcal{D}'$ from $\mathcal{D}$, which yields 5,954 preference pairs with length instructions. The original dataset $\mathcal{D}$ consists of 223 pairs where the two responses have less than $T=10$ words difference, 1,083 pairs where chosen responses are shorter than loser responses, and 1894 pairs where chosen responses are longer. As a result, $\mathcal{D}'$ contains 1,083 pairs where the original winning response loses due to violations of length limits.
We train on $\mathcal{D} \cup \mathcal{D}'$ with the DPO loss, which we call \emph{LIFT-DPO}.

\subsection{Training Details}
In our experiments, we use two sets of base models: Llama2-70B-Base and Llama2-70B-Chat models \citep{touvron2023llama2} and Llama3-8B-Base and Llama3-8B-Instruct. Our DPO training sweeps over a range of learning rates $5e^{-7}$ to $5e^{-6}$ with a cosine learning rate schedule, a batch size of 16, and a dropout rate of 0.1. Specifically for DPO training, we employed a $\beta$ value of 0.1.
For R-DPO, we set $\alpha \in [0.01, 0.1]$\footnote{We had to reverse the sign of the regularization term in Eq. 9 of \citet{park2024disentangling}.}. All Llama2 models are trained for up to 2,000 steps and Llama3 models for up to 20 epochs, and we perform checkpoint selection for early stopping, see \autoref{sec:checkpoint} for more details.

\subsection{Evaluation Method}

We evaluate our models' length instruction-following capabilities on AlpacaEval-LI and MT-Bench-LI, described in \autoref{sec:alpaca_leneval} and \autoref{sec:mtbench_leneval}, as well as general instruction-following on the standard AlpacaEval 2 and MT-Bench benchmarks without length instructions.

For AlpacaEval-LI and MT-Bench-LI, we use the same setup as in AlpacaEval 2 with GPT4 acting as a judge to measure pairwise winrates. 

\begin{table*}[ht]
 \small
\setlength{\tabcolsep}{6.3pt}
  \centering
\begin{tabular}{lrrrcccc}
\toprule
& \multicolumn{3}{c}{AlpacaEval-LI}  &   \multicolumn{3}{c}{MT-Bench-LI} \\
\cmidrule(lr){2-4} \cmidrule(lr){5-7}
 & \multicolumn{1}{c}{Vlt(\%)}   & \multicolumn{1}{c}{Win(\%)} & \multicolumn{1}{c}{Words} & \multicolumn{1}{c}{Vlt(\%)}   & \multicolumn{1}{c}{Win(\%)} & \multicolumn{1}{c}{Words} \\
\midrule
{Llama2-70B-Base (zero shot)}    
& 62.6 & 0.6  & 582& 63.8 & 1.7 & 293 \\
{Llama2-70B-Base + DPO}   
 & 65.8 & 4.6 & 216 & 60.8 & 5.0 & 199 \\
{Llama2-70B-Base + R-DPO \cite{park2024disentangling} ($\alpha=0.01$)} 
& 63.8 & 5.2 & 217 & 57.9 & 2.1 & 194  \\ 
{Llama2-70B-Base + R-DPO \cite{park2024disentangling} ($\alpha=0.1$)}   
& 45.0 & 7.7 & 178 & 39.4 & 8.5 & 161 \\ 
{Llama2-70B-Base + LIFT-DPO} 
& \bf 7.1 & \bf 13.6 & 151 & \bf 10.0& \bf 11.0 &	146
\\
\midrule
{Llama2-70B-Chat} 
& 28.2 & 11.3  & 162 & 38.3 & 11.9 & 168\\
 {Llama2-70B-Chat + DPO} 
 & 15.1 & 10.4 & 135 & 24.2 & 10.8 & 147 \\
{Llama2-70B-Chat + LIFT-DPO} & \bf 2.7 & \bf 14.2 &  140 & \bf 6.7 & \bf 12.5 & 135\\
\bottomrule
\end{tabular}
\caption{{\bf Llama 2 Length Instruction-Following results on the AlpacaEval-LI + MT-Bench-LI benchmarks.} LIFT-DPO yields improved winrates (Win(\%)) and  lower  length instruction following violation rates (Vlt(\%)).}
  \label{tab:leneval-bench-llama2}
\end{table*}

\begin{table*}[ht]
 \small
\setlength{\tabcolsep}{13pt}

  \centering
\begin{tabular}{lrrrcccc}
\toprule
& \multicolumn{3}{c}{AlpacaEval-LI}  &   \multicolumn{3}{c}{MT-Bench-LI} \\
\cmidrule(lr){2-4} \cmidrule(lr){5-7}
 & \multicolumn{1}{c}{Vlt(\%)}   & \multicolumn{1}{c}{Win(\%)} & \multicolumn{1}{c}{Words} & \multicolumn{1}{c}{Vlt(\%)}   & \multicolumn{1}{c}{Win(\%)} & \multicolumn{1}{c}{Words} \\
\midrule 

{Llama3-8B-Base (zero shot)}
& 71.6 &  0.2 & 1935 & 27.9 & 2.3 & 185 \\
{Llama3-8B-Base + DPO}    
& 58.1 & 5.0 & 202 &50.8 &	7.7 &	191 \\
{Llama3-8B-Base + LIFT-DPO}     
& \bf 6.1 & \bf 11.1 & 153 & \bf 13.8 &	\bf 12.9 &	152 \\
\midrule
{Llama3-8B-Instruct} 
& 7.0 & 22.5 &  145 & 20.0 & 20.0 & 140 \\
{Llama3-8B-Instruct + DPO}     
&  7.1 & 25.1 & 143 & 21.3 & 20.0 & 142 \\
{Llama3-8B-Instruct + LIFT-DPO}     
& \bf 3.1 & \bf 25.6 & 161 & \bf 10.8 & \bf 26.3 & 157   \\
\bottomrule
\end{tabular}
 \caption{{\bf
    Llama 3 Length Instruction-Following results on the AlpacaEval-LI + MT-Bench-LI benchmarks.} LIFT-DPO yields improved winrates (Win(\%)) and  lower  length instruction following violation rates (Vlt(\%)).
    }
  \label{tab:leneval-bench-llama3}
\end{table*}

\section{Experimental Results}
\label{sec:results}

We report AlpacaEval-LI winrates (Win(\%)) and violation rates (Vlt\%) for existing SOTA LLMs in 
\autoref{tab:mt-bench-sota}, and for our training variants of Llama2-70B  in \autoref{tab:leneval-bench-llama2}  and Llama3-8B in \autoref{tab:leneval-bench-llama3}. 


Our findings lead to several key observations.

\paragraph{SOTA LLMs fail to follow length instructions}
As demonstrated in \autoref{tab:mt-bench-sota}, state-of-the-art models, such as the GPT-4 series, exhibit significant challenges in adhering to length instructions. Specifically, the latest GPT-4 model (0409) shows a high violation rate of 49.3\% on our AlpacaEval-LI and 44.2\% on MT-Bench-LI. In contrast, the Llama-3 instruct model series displays considerably lower violation rates. For instance, the Llama3-8B-instruct model achieves a violation rate of 7.0\% on AlpacaEval-LI and 20.0\% on MT-Bench-LI, but nevertherless has a lower winrate due to being a less powerful model.

\paragraph{LIFT-DPO models perform well on AlpacaEval-LI and MT-Bench-LI}
\autoref{tab:leneval-bench-llama2} illustrates the effectiveness of our LIFT-DPO training for Llama2 70B models, demonstrating a significant reduction in violation rates compared to both the baseline model and (standard) DPO-trained counterparts. Specifically, the Llama-2-70B-Base model, when subjected to standard DPO training, exhibits a violation rate of 65.8\% on AlpacaEval-LI. 
However, with our LIFT-DPO training, this rate  decreases dramatically to 7.1\%, simultaneously improving the win rate from 4.6\% to 13.6\%. Similarly, for the Llama-2-70B-Chat model, standard DPO results in a violation rate of 15.1\%, whereas our LIFT-DPO training reduces this rate to 2.7\%, and enhances the win rate from 10.4\% to 14.2\%.

On MT-Bench-LI, the Llama-2-70B-Base model has a violation rate of 60.8\% using standard DPO training, which is reduced to 10.0\% with LIFT-DPO, also boosting the win rate from 5.0\% to 11.0\%. For the Llama-2-70B-Chat model, the violation rate decreases from 24.2\% using standard DPO to 6.7\% with LIFT-DPO, with an improvement in the win rate from 10.8\% to 12.5\%.

While the R-DPO baseline improves over standard DPO on both benchmarks especially for a higher $\alpha$ value, it still shows significantly higher violation rates compared to LIFT-DPO, which negatively affects R-DPO's win rates.

\paragraph{LIFT-DPO models show no performance degradation on standard AlpacaEval 2}
We further assessed our LIFT-DPO models using the standard AlpacaEval 2 benchmark, where no length instructions were added and only the original prompts from AlpacaEval 2 were utilized. The results, detailed in Appendix \autoref{tab:alpaca-eval}, indicate no performance degradation when compared to the baselines. Specifically, the Llama-2-70B-Base model achieved a win rate of 8.6\% using standard DPO training, which increased to 9.9\% with our LIFT-DPO training. For the Llama-2-70B-Chat model, the win rates improved from 12.6\% using DPO to 12.9\% with LIFT-DPO. However, the Llama-3-8B-Base models yielded a slight decrease in winrate from 7.8\% with DPO to 7.2\% with LIFT-DPO, although the LC (length-controlled)-winrate actually increased from 13.9\% to 15.7\% (as the average response length decreased). Similarly, the Llama-2-8B-Instruct models have a winrate of 25.8\% with DPO, which slightly decreased to 22.7\% with our LIFT-DPO training, although the LC-winrate actually increased from 26.3\% to 26.5\%.  In summary, our LIFT-DPO models exhibit comparable performance to standard DPO when length instructions are not applied.
We observed similar results on standard MT-Bench as shown in Appendix \autoref{tab:mt-bench-single}.

\begin{figure}[t!]
    \centering
    \includegraphics[width=\columnwidth]{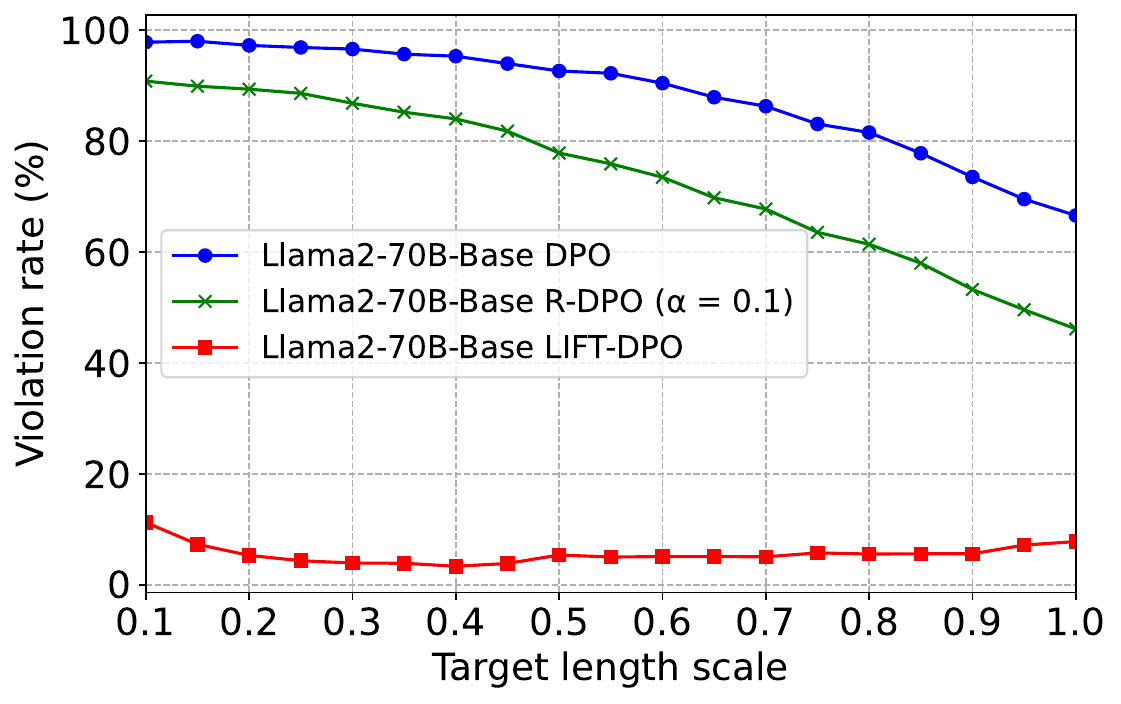}
    \caption{The violation rate of DPO or R-DPO Llama2-70B models on AlpacaEval-LI increases as the target length shortens (via target length scale). However, LIFT-DPO consistently follows length instructions,  maintaining a low violation rate independent of length scale.}    \label{fig:clif_vs_normal_dpo_violation_rate}
\end{figure}

\paragraph{LIFT-DPO can follow out-of-distribution length instructions better than existing methods}
To increase the difficulty of our AlpacaEval-LI benchmark, we can progressively decrease the limit in the length instructions  by applying a scaling factor to the existing values ranging from 0.9 down to 0.1. This adjustment introduces a spectrum of challenging length constraints. We assessed the performance of various models based on Llama-2-70B-Base, including  standard DPO, R-DPO and LIFT-DPO,  and plotted their violation rates. The results are provided in  \autoref{fig:clif_vs_normal_dpo_violation_rate}.
The analysis reveals that the standard DPO model exhibits increasingly higher violation rates as the length scale decreases, with rates escalating from below 50\% to almost 100\% when the scale factor is set to 0.1. This indicates significant difficulties in adhering to stringent length constraints for this model. The R-DPO model displays trends similar to standard DPO, suggesting that while it can reduce the generation length, it lacks the capability to precisely steer it. In contrast, our LIFT-DPO model consistently maintains a low violation rate (below 10\%) across all tested length scales.  We observe similar trends on MT-Bench-LI, see Appendix \autoref{fig:clif_vs_normal_dpo_violation_rate_mtbench} for details.

\begin{figure}
    \centering
    \includegraphics[width=\columnwidth]{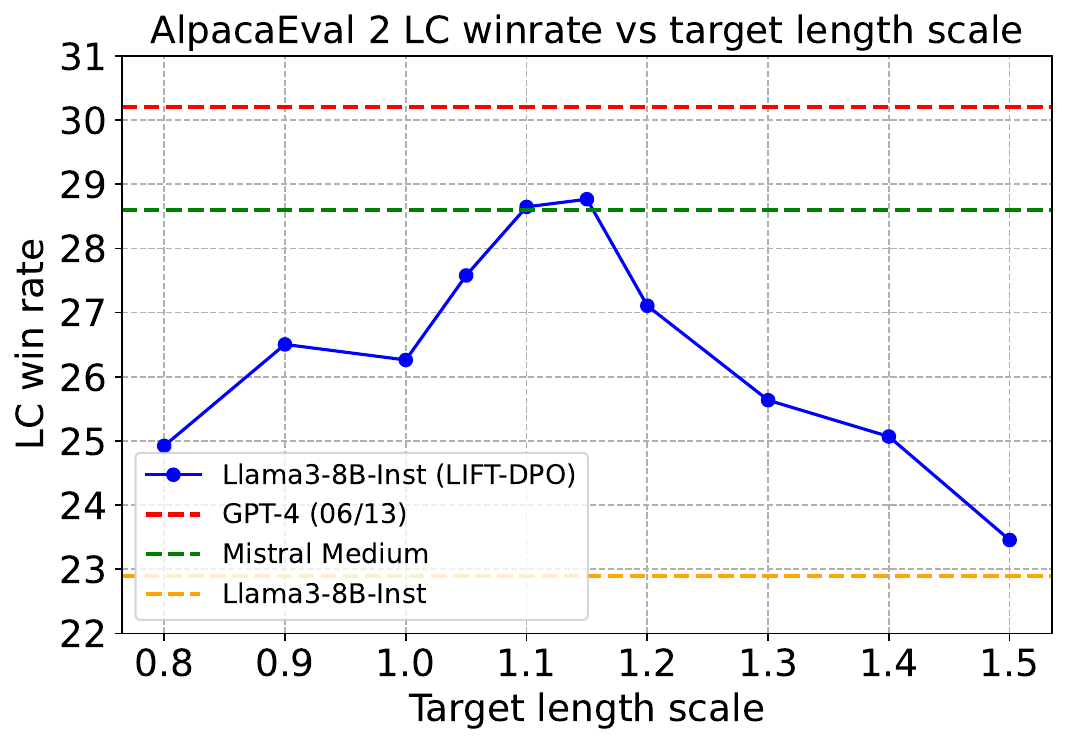}
    \caption{AlpacaEval 2 Length-Controlled (LC) winrate vs target length scale. We show that our LIFT-DPO Llama-3-8B-Instruct model can be controlled to produce different length responses, which affects overall LC winrate. The optimum length choice produces scores similar to Mistral Medium, and is superior to the original Llama-3-8B-Instruct model.
    }
    \label{fig:lc_winrate_vs_scale}
\end{figure}

\paragraph{Robustness of Length Controlled AlpacaEval}

Previous research has acknowledged the presence of length bias, and designers have introduced measures to mitigate it, notably through Length-Controlled (LC) AlpacaEval, which incorporates an LC winrate that considers generation length~\cite{dubois2024length}. Despite these efforts, we find that the LC winrate can still be manipulated by adjusting the length instructions. By scaling the length constraints as we did in AlpacaEval-LI and measuring the AlpacaEval LC winrate, we observe significant fluctuations in the results, as shown in \autoref{fig:lc_winrate_vs_scale}. The LC winrate varies dramatically, from 23\% up to 29\%. 
In contrast, in our work we argue that expected length is ill-defined in many queries (see motivation in  \autoref{sec:intro}), and that length instruction evaluation helps remove this ambiguity, and hence also any potential gameability. 




\section{Conclusion}

To address the issue of length bias in general instruction following,
 we propose length instructions, 
 which assess models' abilities to generate responses 
 within given length limits.
We introduce two Length-Instructed (LI) benchmarks,  MT-Bench-LI and  AlpacaEval-LI, and
show that SOTA models surprisingly fail to follow length instructions on these benchmarks. 
 We hence propose Length-Instruction Fine-Tuning (LIFT), a method that augments existing general instruction-following examples with varying length limits. LIFT-DPO models show significant improvement in their ability to control output length while maintaining high response quality. 
Our length instruction following approach provides a way to compare models without length bias, as it does not suffer from the gameability of
simply increasing model response length, as that leads to a violation. 
In addition, augmenting general instructions with length limits allows for more controllability for users in real-world use cases.

\section{Limitations}
In this paper, the length limit is set in terms of the number of words, but more generally it can be set in number of characters, or some other measure.
Another direction of generalization can be allowing length instructions to be phrased using different wording instead of a fixed template, so users can specify the limit in their own words, such as ``Keep the response under 100 words.''. 
We also did not address other kinds of length instructions such as ``write at least 100 words''.
While this paper attempts to address length bias in model evaluations through length instructions, this bias may also arise from a natural human preference for longer and more detailed responses. 
Future research could further explore human desired response lengths across different instructions. Such studies could further enhance the alignment of models with human expectations.
Another possible cause of longer responses could be related to the increased computation allowance that comes with more tokens, which can benefit from future analysis.

\bibliography{custom}

\appendix

\section{Word Count Function We Use}
\label{sec:appex_wordcount}
\begin{lstlisting}
from nltk.tokenize import word_tokenize
import string

def count_words(text) -> int:
    # Count the number of words
    # while excluding punctuations
    return len([word for word in word_tokenize(text) if word not in string.punctuation])
\end{lstlisting}


\section{Additional Results on SOTA models' length following measurements}
\label{sec:appex_sota_len_following_claude3_mistral}
\begin{figure*}
    \includegraphics[width=0.48\linewidth]{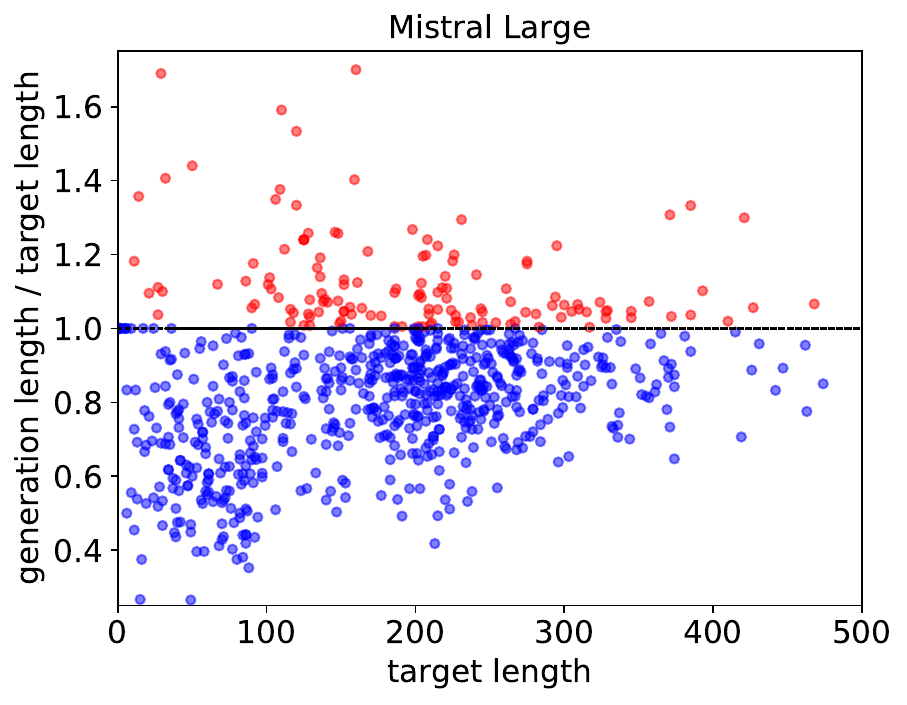}
    \hfill
    \includegraphics[width=0.48\linewidth]{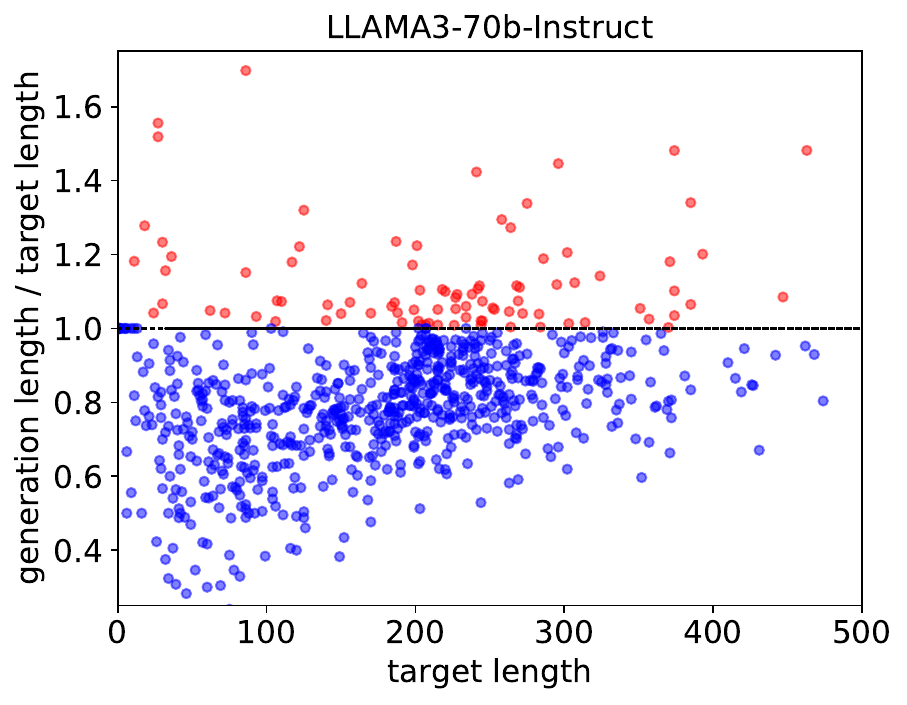}
    \caption{The length instruction following ability of Mistral Large and LLAMA3-70b-Instruct on 802 AlpacaEval Length-Instructed (LI) examples. The scatter plots display each sample from the AlpacaEval LI dataset, with the target length plotted on the x-axis and the ratio of the actual generated length to the target length on the y-axis. Red dots represent violations where the generated length exceeds the target limit, while blue dots satisfy the limit.
    }
    \label{fig:scatter_claude3_mistral}
\end{figure*}

We plot the generation lengths over target instruction lengths on AlpacaEval-LI for Mistral Large and LLAMA3-70b-Instruct in \autoref{fig:scatter_claude3_mistral}. The scatter plots reveal that both models occasionally fail to meet the length constraints.

\section{Training and test length distribution}
\begin{figure*}
    \includegraphics[width=0.495\linewidth]{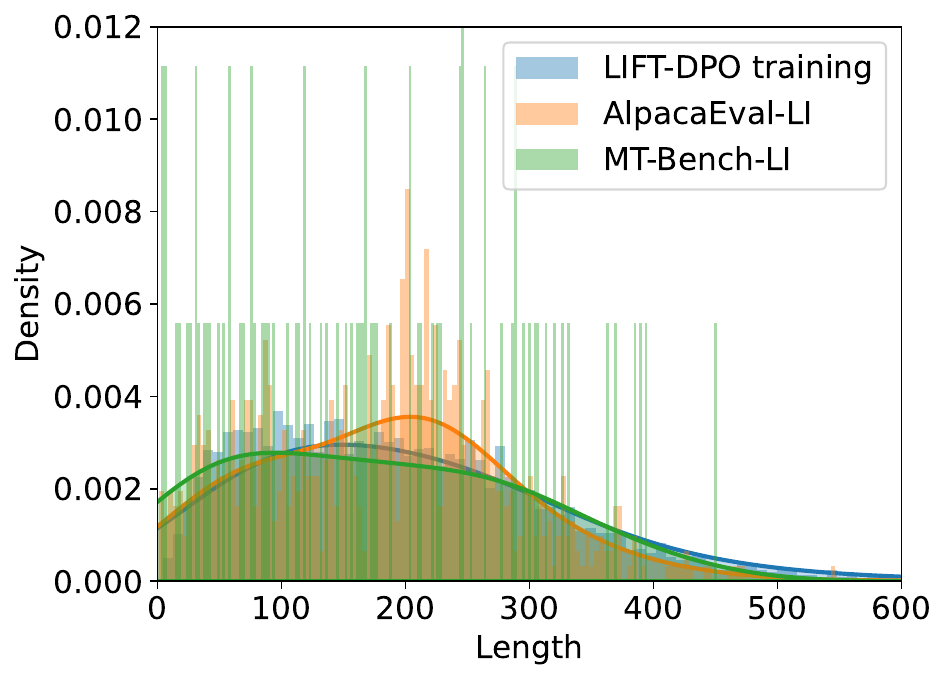}
    \hfill
    \includegraphics[width=0.48\linewidth]{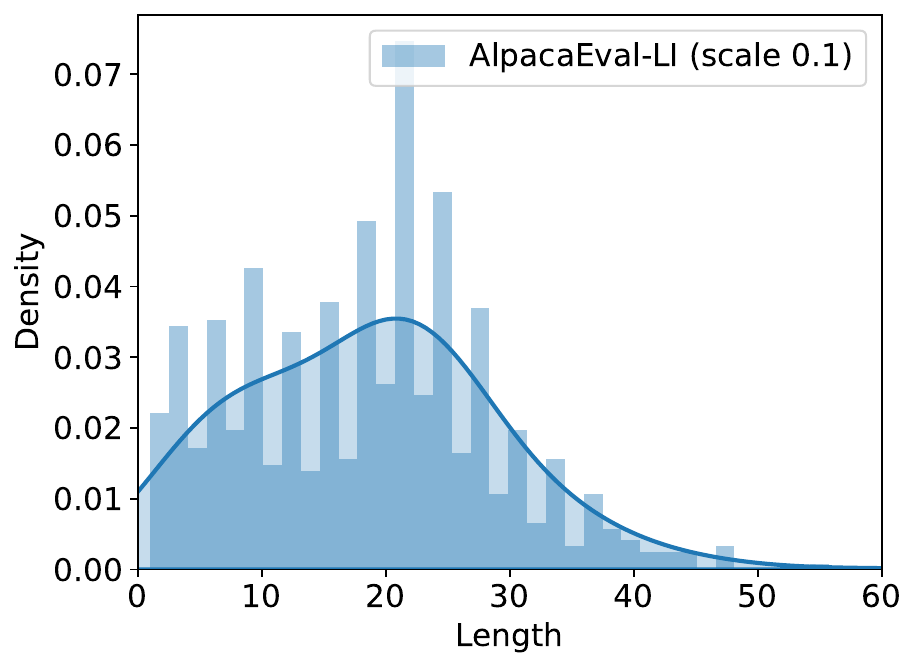}
    \caption{The distribution of length constraints across LIFT-DPO training data, AlpacaEval-LI, and MT-Bench-LI. Additionally, we also include a plot of the AlpacaEval-LI length constraints scaled by a factor of 0.1.
    }
    \label{fig:train_test_dist}
\end{figure*}

\autoref{fig:train_test_dist} illustrates the distribution of length constraints in our LIFT-DPO training data alongside those in AlpacaEval-LI and MT-Bench-LI. We observed that the majority of our training data features length constraints ranging from 50 to 300, a range that is consistent with that of AlpacaEval-LI. Additionally, we have depicted the distribution of length constraints in AlpacaEval-LI scaled by a factor of 0.1 in \autoref{fig:train_test_dist}. Nearly all scaled length constraints fall below 50, constituting only a small fraction of the length constraints present in our training dataset.

\section{Checkpoint Selection}
\label{sec:checkpoint}
We perform checkpoint selection by saving a checkpoint every 200 steps and at the end of each epoch. We then evaluate these checkpoints using GPT-4-Turbo on a set of 253 validation examples, which are derived from various sources as outlined by \citet{li2023self}. The LI (Length-Instructed) validation set is augmented from the same validation set but includes length limits, using the minimum length from three strong LLMs in \autoref{sec:target-choice}.

For the standard instruction-following validation set, each new model checkpoint is evaluated by comparing its generations pairwise with those from the previous checkpoint, utilizing the AlpacaEval evaluation prompt format \citep{alpaca_eval}. For length-instructed tasks, evaluations are conducted pairwise against a baseline from one of the three LLMs, specifically the one whose generation length matches the length limit specified in the prompt. The win rate of a model checkpoint is calculated as the average of the win rates on both the instruction-following validation set and the LI validation set. We implement early stopping if we observe a decrease in this average win rate.

\section{MT-Bench Results}
In the standard MT-Bench evaluation, models employ different temperatures (including 0) for different categories during inference time. To expand the size of MT-Bench-LI via sampling, we standardized the temperature setting to 0.7 across all categories for pairwise baseline models as well as models being tested. However, for the standard MT-Bench evaluation reported in \autoref{tab:mt-bench-single}, we switch back to the original setup using different temperatures for different categories and assessing performance on 80 unique questions.

\section{Decoding Parameters}
During inference time, except for the standard MT-Bench evaluations, we apply consistent hyperparameter settings for the Llama models. For the Llama2 models, we set the temperature to 0.7, with a maximum token limit of 2048. For the Llama3 models, the temperature is adjusted to 0.6, maintaining the same top-p of 0.9, but with an increased maximum token limit of 4096. We  consistently set top-p to 0.9 for AlpacaEval 2 and AlpacaEval-LI and top-p to 1.0 for MT-Bench and MT-Bench-LI.

\section{Additional Length Instruction Following Results}
\begin{figure}[t]
    \centering
    \includegraphics[width=\columnwidth]{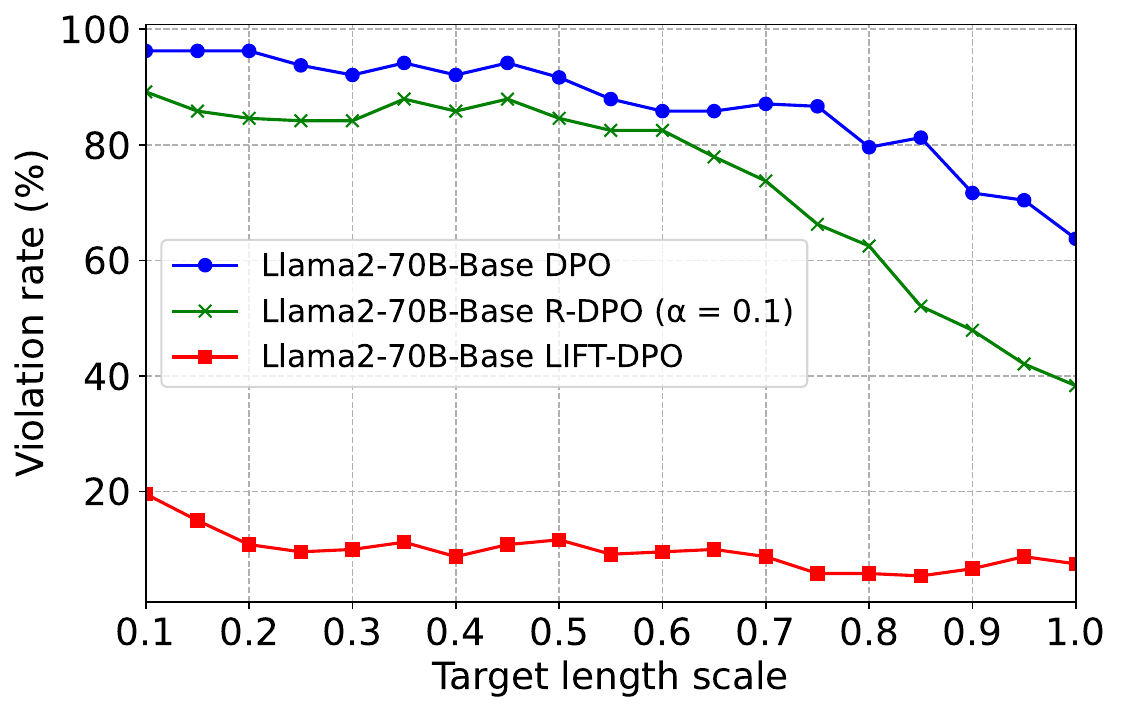}
    \caption{The violation rates of DPO, R-DPO, LIFT-DPO trained models based on Llama2-70B models on MT-Bench-LI as the target length shortens (via target length scale).}
    \label{fig:clif_vs_normal_dpo_violation_rate_mtbench}
\end{figure}

\begin{figure}[t]
    \centering
    \includegraphics[width=\columnwidth]{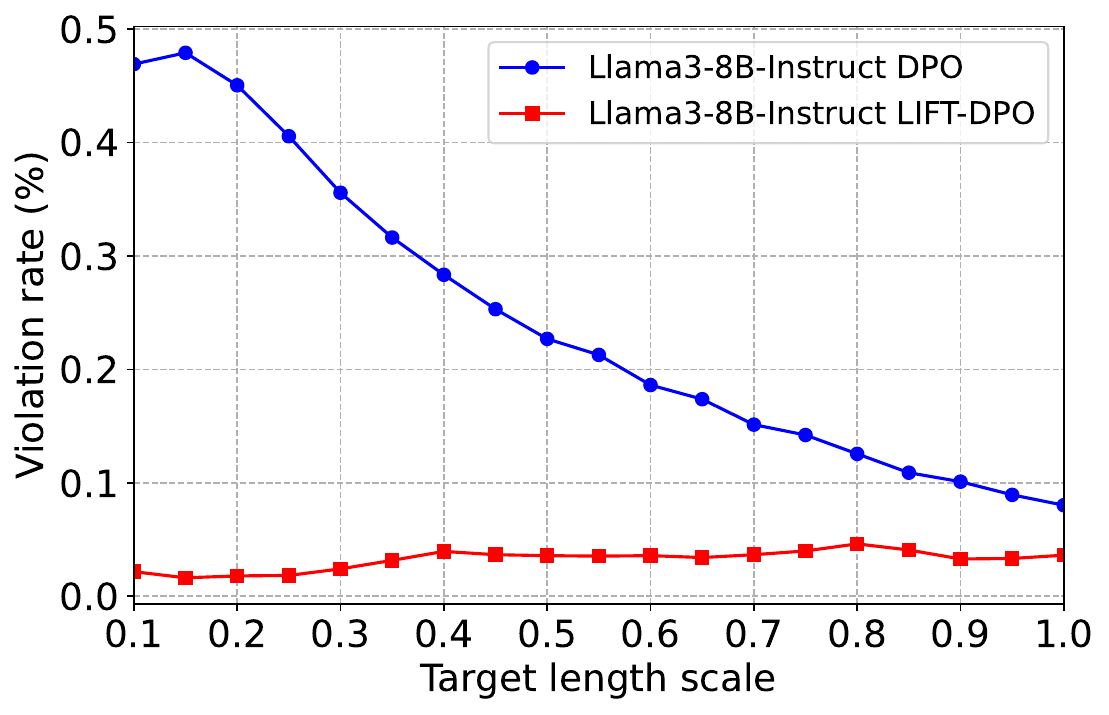}
    \caption{The violation rates of DPO, LIFT-DPO trained models based on Llama3-8B-Instruct on AlpacaEval-LI as the target length shortens (via target length scale).}
    \label{fig:clif_vs_normal_dpo_violation_rate_alpacaeval_llama3_instruct}
\end{figure}

In our MT-Bench-LI evaluations, we progressively reduced the length instructions by applying scaling factors to the existing values, ranging from 0.9 down to 0.1. We assessed the performance of various models based on the Llama-2-70B-Base, including standard DPO, R-DPO, and LIFT-DPO, and plotted their violation rates as shown in \autoref{fig:clif_vs_normal_dpo_violation_rate_mtbench}). The results indicate that our LIFT-DPO trained model significantly outperforms both DPO and R-DPO in adhering to length constraints. Specifically, the LIFT-DPO model maintains a violation rate below 20\% across all scaling factors, whereas both DPO and R-DPO models exhibit violation rates exceeding 80\% when the scaling factor is reduced to less than 0.6.
Additionally, we analyzed the performance of models based on Llama-3-8B-Instruct on AlpacaEval-LI under gradually reduced length limits. The observed trend is similar to that of MT-Bench-LI, as depicted in \autoref{fig:clif_vs_normal_dpo_violation_rate_alpacaeval_llama3_instruct}.

\section{AlpacaEval Results \& MT-Bench Results}
The results of the LIFT-DPO models on standard AlpacaEval and MT-Bench are detailed in \autoref{tab:alpaca-eval} and \autoref{tab:mt-bench-single}, respectively. Our analysis reveals that the LIFT-DPO models exhibit no performance degradation when compared to the standard DPO models on these benchmarks.

\begin{table*}[ht]
 \small
\setlength{\tabcolsep}{20pt}

  \centering
\begin{tabular}{lccccc}
\toprule
\emph{Standard models}  & Vlt(\%)  &  LC-Win(\%)  & Win(\%)  & Words\\
\midrule
{GPT4 Turbo(1106-preview)} & 91.1 & 50 & 50 & 324 \\
{GPT4 Turbo(0409-preview)}  & 77.1 & 55.0 & 46.1 & 277 \\
{GPT4 Omni}   & 77.8 &  57.5 & 51.3 & 282 \\
{Claude 3 Opus (02/29}  & 57.8  & 40.5 & 29.1 & 219\\
{Mistral Large (24/02)} & 49.7 & 32.7 & 21.4 & 223 \\
{Llama2-70B Chat}  & 84.8 & 13.9 & 14.7 & 296 \\
{Llama3-70B Instruct} & 84.2   & 34.4   & 33.2&  302 \\
{Llama3-8B Instruct}  & 88.6 & 22.9 & 22.6 & 303 \\
\midrule \midrule
\emph{Llama2-70B Models} \\ 
\midrule
{Llama2-70B + DPO}  
    & 60.7 &  13.1 & 8.6 &  211 \\
{Llama2-70B + LIFT-DPO}  
& 65.7 & 15.4  & 9.9 & 220\\
{Llama2-70B + R-DPO ($\alpha=0.01$)} & 57.9 & 11.3  & 7.5 & 204  \\ 
{Llama2-70B + R-DPO ($\alpha=0.1$)}    & 48.6  & 13.6 & 8.0  & 187 \\ 
\midrule
{Llama2-70B-Chat + DPO} & 66.8& 23.3 & 12.6 & 218 
\\
{Llama2-70B-Chat + LIFT-DPO}   &  75.9 & 20.5 & 12.9 & 242 \\
\midrule    \midrule    
\emph{Llama3-8B Models} \\ 
\midrule
{Llama3-8B + DPO}   & 45.1  & 13.9 &  7.8  & 188 \\
{Llama3-8B + LIFT-DPO}  & 33.9 & 15.7  & 7.2 & 158 \\
\midrule
{Llama3-8B-Instruct + DPO} & 86.5 & 26.3 &  25.8 & 308 \\
{Llama3-8B-Instruct + LIFT-DPO} & 85.1 & 26.5 & 22.7 & 285 \\
\bottomrule
\end{tabular}
    \caption{{\bf Results on the AlpacaEval benchmark}. LIFT-DPO still maintains good performance in the standard (non-length) instruction-following setup.  
    }
  \label{tab:alpaca-eval}
\end{table*}

\begin{table*}[ht]
 \small
\setlength{\tabcolsep}{12pt}

  \centering
\begin{tabular}{lcccc}
\toprule
\emph{Llama2 Models}  & Overall  & Math, Coding & Humanities, Extraction  & \multirow{2}{*}{Words} \\
& Score & \& Reasoning & \& STEM, Roleplay, Writing  & \\
\midrule
\midrule
{Llama2-70B + DPO}  
  & 7.45   & 5.30 & 8.74 & 189  \\
{Llama2-70B + LIFT-DPO}   & 7.54  & 4.77 & 9.21  & 275 \\
{Llama2-70B + R-DPO ($\alpha=0.01$)} & 6.65 & 4.17 & 8.14  & 181   \\ 
{Llama2-70B + R-DPO ($\alpha=0.1$)}  & 6.53   & 3.53 & 8.33 & 163   \\ 
\midrule
{Llama2-70B-Chat + DPO} & 7.58  & 5.03 & 9.10 & 218 
\\
{Llama2-70B-Chat + LIFT-DPO}  & 7.45  & 4.70  & 9.10 & 213  \\
\midrule    \midrule    
\emph{Llama3-8B Models} \\ 
\midrule
{Llama3-8B + DPO}  & 7.11 &  8.44 & 4.90  &  158 \\
{Llama3-8B + LIFT-DPO} &  6.99 & 8.54 & 4.40 & 138  \\
\midrule
{Llama3-8B-Instruct + DPO} & 8.38 & 6.30 & 9.62 & 263  \\
{Llama3-8B-Instruct + LIFT-DPO} & 8.32 & 6.27 & 9.55 & 237   \\
\bottomrule
\end{tabular}
    \caption{{\bf Results on the MT-Bench benchmark}. LIFT-DPO still maintains good performance in the standard (non-length) instruction-following setup.  
    }
  \label{tab:mt-bench-single}
\end{table*}

\end{document}